\documentclass{sig-alternate-05-2015}

\usepackage[lined,linesnumbered,figure]{algorithm2e}
\usepackage{wrapfig}
\usepackage{verbatim}
\usepackage{graphicx}
\usepackage{amsmath,amsfonts,amssymb}
\usepackage{caption}
\usepackage[draft]{hyperref}
\newcommand{\bu}{\mathbf{u}}
\newcommand{\bdu}{\mathbf{\triangle u}}
\newcommand{\bv}{\mathbf{v}}
\newcommand{\bdv}{\mathbf{\triangle v}}
\newcommand{\grad}{\vec\nabla}

\begin{document}

\title{Network--Efficient Distributed Word2vec \\ Training System for Large Vocabularies}
\numberofauthors{1}
\author{
\begin{tabular}[t]{p{\textwidth}}\centering
 Erik Ordentlich, Lee Yang, Andy Feng, Peter Cnudde, Mihajlo Grbovic \\
 \affaddr{Yahoo, Inc.} \\
 \{eord, leewyang, afeng, pcnudde, mihajlo\}@yahoo-inc.com \\
 \end{tabular} \\
 \begin{tabular}[t]{cc}
   Nemanja Djuric\titlenote{Work done while with Yahoo, Inc.}, Vladan Radosavljevic\raisebox{9pt}{$\ast$} & Gavin Owens\raisebox{9pt}{$\ast$} \\
   \affaddr{Uber ATC} & \affaddr{Deco Software}
 \end{tabular}
}

\maketitle
\begin{abstract}
Word2vec is a popular family of algorithms for unsupervised training of dense vector representations of words on large text corpuses. The resulting vectors have been shown to capture semantic relationships among their corresponding words, and have shown promise in reducing a number of natural language processing (NLP) tasks to mathematical operations on these vectors. While heretofore applications of word2vec have centered around vocabularies with a few million words, wherein the vocabulary is the set of words for which vectors are simultaneously trained, novel applications are emerging in areas outside of NLP with vocabularies comprising several 100 million words. Existing word2vec training systems are impractical for training such large vocabularies as they either require that the vectors of all vocabulary words be stored in the memory of a single server or suffer unacceptable training latency due to massive network data transfer. In this paper, we present a novel distributed, parallel training system that enables unprecedented practical training of vectors for vocabularies with several 100 million words on a shared cluster of commodity servers, using far less network traffic than the existing solutions. We evaluate the proposed system on a benchmark dataset, showing that the quality of vectors does not degrade relative to non-distributed training. Finally, for several quarters, the system has been deployed for the purpose of matching queries to ads in Gemini, the sponsored search advertising platform at Yahoo, resulting in significant improvement of business metrics. 
\end{abstract}

\section{Introduction}
\begin{figure*}
\footnotesize
  \begin{center}
  \fbox{
    \parbox{\textwidth}{ {\bf gas\_caps}  {\bf gas\_cap\_replacement\_for\_car} slc\_679f037d {\bf gas\_door\_replacement\_for\_car} slc\_466145af1 {\bf fuel\_door\_covers} adid\_28540536 slc\_348709d7 {\bf autozone\_auto\_parts} adid\_33183157 {\bf auoto\_zone} slc\_8dcdab5d slc\_58f979b6 \\

{\bf hoka\_running\_shoe\_reviews} adid\_22830771 {\bf hoka\_shoes\_for\_bad\_feet}  {\bf hoka\_shoes\_amazon} slc\_231g5a94 {\bf zappos\_shoes} slc\_7c126f71 {\bf hoka\_walking\_shoes} \\

{\bf king\_tut } {\bf king\_tut\_exhibit }  {\bf king\_tut\_exhibit\_seattle\_2015} slc\_726y6j51 {\bf charlies\_seattle} adid\_55774014

  }}
  \end{center}
  \caption{Snippet from large training corpus for sponsored search application.}
\label{fig:q2atrainingex}
\end{figure*}

Embedding words in a common vector space can enable machine learning algorithms to achieve better performance in natural language processing (NLP) tasks. Word2vec \cite{word2vec} is a recently proposed family of algorithms for training such vector representations from unstructured text data via shallow neural networks. The geometry of the resulting vectors was shown in~\cite{word2vec} to capture word semantic similarity through the cosine similarity of the corresponding vectors as well as more complex semantic relationships through vector differences, such as vec(``Madrid'') - vec(``Spain'') + vec(``France'') $\approx$ vec(``Paris'').

More recently, novel applications of word2vec involving unconventional generalized ``words'' and training corpuses have been proposed. These powerful ideas from the NLP community have been adapted by researchers from other domains to tasks beyond representation of words, including relational entities \cite{bordes2013translating,socher2013reasoning}, general text-based attributes \cite{kiros2014multiplicative}, descriptive text of images \cite{kiros2013multimodal}, nodes in graph structure of networks \cite{perozzi2014deepwalk}, and queries \cite{query2vec}, to name a few.

While most NLP applications of word2vec do not require training of large vocabularies, many of the above mentioned real-world applications do. For example, the number of unique nodes in a social network \cite{perozzi2014deepwalk} or the number of unique queries in a search engine \cite{query2vec} can easily reach few hundred million, a scale that is not achievable using existing word2vec implementations.

The training of vectors for such large vocabularies presents several challenges.  In word2vec, each vocabulary word has two associated $d$-dimensional vectors which must be trained, respectively referred to as input and output vectors, each of which is represented as an array of $d$ single precision floating point numbers~\cite{word2vec}.   To achieve acceptable training latency, all vectors need to be kept in physical memory during training, and, as a result, word2vec requires $2\cdot d\cdot 4\cdot |{\cal V}|$ bytes of RAM to train a vocabulary ${\cal V}$.  For example, in Section~\ref{sec:applications},  we discuss the search advertisement use case with 200 million generalized words and $ d = 300 $ which would thus require $ 2\cdot 300\cdot 4\cdot 200M $ =  480GB memory which is well beyond the capacity of typical commodity servers today.  Another issue with large vocabulary word2vec training is that the training corpuses required for learning meaningful vectors for such large vocabularies, are themselves very large, on the order of 30 to 90 billion generalized words in the mentioned search advertising application, for example, leading to potentially prohibitively long training times.  This is problematic for the envisioned applications which require frequent retraining of vectors as additional data containing new ``words'' becomes available.  The best known approach for refreshing vectors is to periodically retrain on a suitably large window comprised of the most recent available data.  In particular, we found that tricks like freezing the vectors for previously trained words don't work as well.  The training latency is thus directly linked to staleness of the vectors and should be kept as small as feasible without compromising quality.

Our main contribution is a novel distributed word2vec training system for commodity shared compute clusters that addresses these challenges.   The proposed system:
\begin{enumerate}
\setlength{\itemsep}{2pt}
\item allows very large vocabulary sizes by distributing the word vectors in a novel fashion across multiple servers.
\item parallelizes vector training to reduce training latency to practical ranges, enabling frequent retraining to incorporate new data.
\end{enumerate}
As discussed in Section~\ref{sec:previous}, to the best of our knowledge, this is the first word2vec training system that is truly scalable in both of these aspects.\footnote{In this work, we focus exclusively on scaling word2vec.   We leave the suitability and scalability of the more recent ``count'' based embedding algorithms that operate on word pair co-occurrence counts~\cite{levyw2v,glove,swivel} to the data sets and vocabulary sizes of interest here as open questions, noting only that the vocabularies considered in published experiments involving these alternatives is at most 500,000 words.}


We have implemented the proposed word2vec training system in Java and Scala, leveraging the open source building blocks Apache Slider~\cite{slider} and Apache Spark~\cite{spark} running on a Hadoop YARN-scheduled cluster~\cite{hadoop,yarn}.  Our word2vec solution enables the aforementioned applications to efficiently train vectors for unprecedented vocabulary sizes. Since late 2015, it has been incorporated into the Yahoo Gemini Ad Platform (\url{https://gemini.yahoo.com}) as a part of the ``broad'' ad matching pipeline, with regular retraining of vectors based on fresh user search session data.



\section{Sponsored search use case} \label{sec:applications}

Sponsored search is a popular advertising model \cite{jansen2008sponsored} used by web search engines, such as Google, Microsoft, and Yahoo, in which advertisers {\it sponsor} the top web search results in order to redirect user's attention from {\it organic} search results to ads that are highly relevant to the entered query.

Most search engines provide a self-service tool in which the advertisers can create their own ads by providing ad creative to be shown to the users, along with a list of bid terms (i.e., queries for which advertisers wish to show their ad). Due to a large number of unique queries it is challenging for advertisers to identify all queries relevant to their product or service. For this reason search engines often provide a service of ``broad'' matching, which automatically finds additional relevant queries for advertisers to bid on. This is typically implemented by placing queries and ads in a common feature space, such as bag-of-words using tf-idf weighting, and calculating similarity between ads and queries using a feature space metric in order to find good broad match candidates.

In an unconventional application of word2vec to historical search logs, one could train query and ad vectors that capture semantic relationships and find relevant broad match candidates in the resulting feature space. The idea of using word2vec to train query representations is not new and has been suggested by several researchers in the past \cite{mitra2015exploring,query2vec}. However, until now, it was not possible to use the algorithm to its fullest extent due to computational limitations of existing word2vec implementations.


The sponsored search training corpus consists of billions of user search sessions each comprising generalized ``words'' corresponding to entire user queries (not the individual words in the queries), clicked hyperlinks, and clicked advertisements, ordered according to the temporal ordering of the corresponding user actions.  Figure~\ref{fig:q2atrainingex} shows a snippet from such a training corpus wherein the clicked ads and search link clicks are encoded as string IDs prefixed by ``adid\_'' and ``slc\_'', respectively.  The queries are highlighted in bold. 

The goal is to train vector representations for queries, hyperlinks, and advertisements, and to use the semantic similarity captured by these vectors to target advertisements to semantically relevant queries that might otherwise not be found to be relevant using more conventional measures, such as prior clicks or the number of constituent words common to the query and advertisement meta data (i.e., title, description, bid keywords).  Note that although the search result hyperlinks clicked by the user are not needed for the sponsored search system, they are nevertheless important to include during training as they help propagate relevance between the queries and ads of interest.  

Given trained query and ad vectors, finding relevant queries for a given ad amounts to calculating cosine similarity between the ad vector and all query vectors. The $K$ queries with the highest similarity are retrieved as broad matches. 

\begin{figure}
\centering
\includegraphics[height=2.1in]{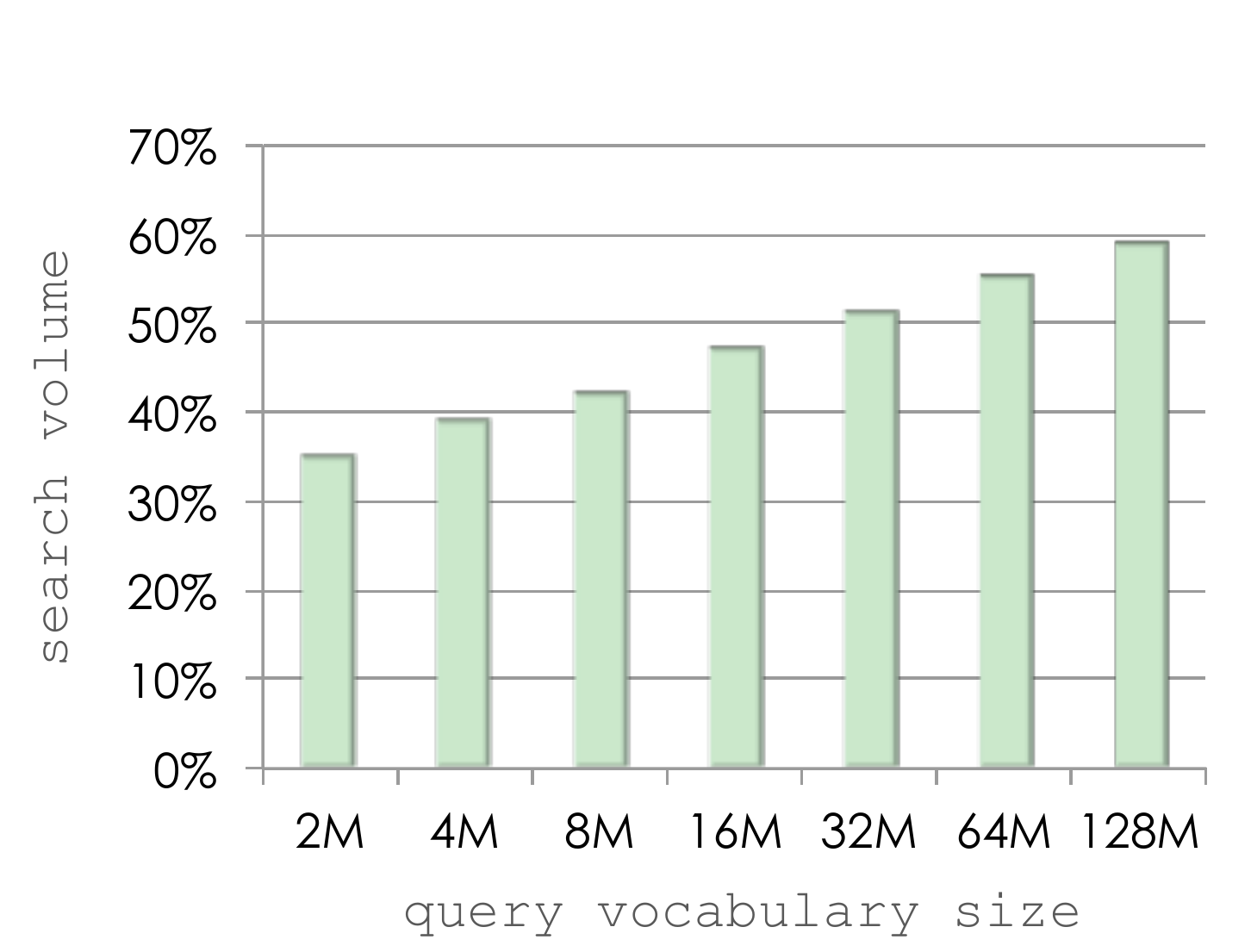}
\caption{Fraction of query occurrences (search volume) vs. number of queries in vocabulary for which vectors have been trained}
\label{fig:coverage}
\end{figure}

As illustrated in Figure~\ref{fig:coverage} for representative search session data, the fraction of query occurrences in the search sessions for which vectors are available, and hence for which potential ads can be found using this vector-based approach, increases at a steady pace with the number of queries in the vocabulary, even with as many as 120 million queries, each occurring at least 5 times.  This observation suggests that this application can benefit greatly from vocabularies of 200 million or more generalized words.  Moreover, we found that there are around 800 million generalized words that occur 5 or more times in our largest data sets, indicating that additional scaling far beyond 200 million is well worth pursuing. 

The results of~\cite{query2vec} were based on training the largest vocabulary that could fit into the large memory of a special purpose server, which resulted in learned vector representations for about 45 million words.  The proposed training system herein enables increasing this by several fold, resulting in far greater coverage of queries and a potentially significant boost in query monetization, as indicated by Figure~\ref{fig:coverage}.  

\section{The word2vec training problem}
\label{sec:word2vecopt}
In this paper we focus on the skipgram approach with random negative examples proposed in~\cite{word2vec}.  This has been found to yield the best results among the proposed variants on a variety of semantic tests of the resulting vectors~\cite{levyw2v,word2vec}.  Given a corpus consisting of a sequence of sentences $s_1,s_2, \ldots, s_n$ each comprising a sequence of words $s_i =  w_{i,1},w_{i,2}, \ldots,w_{i,m_i}$, the objective is to maximize the log likelihood:
\begin{multline}
  \sum_{i=1}^n \; \sum_{j : w_{i,j} \in {\cal V} } \; \sum_{\substack{k \neq j: |k-j| \leq b_{i,j} \\ w_{i,k} \in {\cal V}}} \Bigg[ \log \sigma \big(\bu(w_{i,j})\bv^\mathsf{T}(w_{i,k})\big) + \\
  \sum_{\tilde{w} \in {\cal N}_{i,j,k}}\log \Big(1-\sigma \big(\bu(w_{i,j})\bv^\mathsf{T}(\tilde{w})\big)\Big) \Bigg]
  \label{eq:w2vobjective}
\end{multline}
over input and output word row vectors $ \bu(w) $ and $ \bv(w) $ with $ w $ ranging over the words in the vocabulary $ {\cal V} $, where:
  \begin{itemize}
  \setlength{\itemsep}{2pt}
  \item $ \sigma(\cdot) $ denotes the sigmoid function $ \sigma(x) = 1/(1+e^{-x}) $;
  \item window sizes $  b_{i,j} $ are randomly selected so that each inner sum includes between 1 and a maximum $ B $ terms, as in~\cite{word2vec} and its open--source implementation;\footnote{Throughout, it is assumed that words not in the vocabulary or words omitted due to the subsampling of frequent words, following~\cite{word2vec}, do not count towards window or context size. That is, we assume ``dirty'' contexts using the terminology of~\cite{levyw2v}, consistent with the open--source version of~\cite{word2vec}.}
  \item negative examples $ {\cal N}_{i,j,k} $ associated with positive output word $ w_{i,k} $ are selected randomly according to a probability distribution suggested in~\cite{word2vec};
  \item and the vocabulary $ {\cal V} $ consists of the set of words for which vectors are to be trained.
  \end{itemize}
  
We follow~\cite{word2vec} for setting ${\cal V} $ and select words occurring in the corpus a sufficient number of times (e.g., at least $ 5 $ times), or, if this results in too many words, as the most frequently occurring $ N $ words, where $ N $ is the largest number words that can be handled by available computational resources.  We further also assume a randomized version of~(\ref{eq:w2vobjective}) according to the subsampling technique of~\cite{word2vec}, which removes some occurrences of frequent words.

The algorithm for maximizing~(\ref{eq:w2vobjective}) advocated in~\cite{word2vec}, and implemented in its open--source counterpart, is a minibatch stochastic gradient descent (SGD).  Our training system is also based on minibatch SGD optimization of~(\ref{eq:w2vobjective}), however, as described in Section~\ref{sec:alg}, it is carried out in a distributed fashion in a manner quite different from the implementation of~\cite{word2vec}.  Any form of minibatch SGD optimization of~(\ref{eq:w2vobjective}) involves the computation of dot products and linear combinations between input and output word vectors for all pairs of words occurring within the same window (with indices in $\{k \neq j: |k-j| \leq b_{i,j}\}$).  This is a massive computational task when carried out for multiple iterations over data sets with tens of billions of words, as encountered in applications described in the previous section.

\section{Existing word2vec systems}  \label{sec:previous}
\subsection{Single machine}
Several existing word2vec training systems are limited to running on a single machine, though with multiple parallel threads of execution operating on different segments of training data.  These include the original open source implementation of word2vec~\cite{word2vec}, as well as those of Medallia~\cite{medallia}, and Rehurek~\cite{gensim}.   As mentioned in the introduction, these systems would require far larger memory configurations than available on typical commodity-scale servers.

\subsection{Distributed data-parallel}
A similar drawback applies to distributed data-parallel training systems like those available in Apache Spark MLLib \cite{mllib} and Deeplearning4j \cite{dl4j}.  In the former, in each iteration the Spark driver sends the latest vectors to all Spark executors. Each executor modifies its local copy of vectors based on its partition of the training data set, and the driver then combines local vector modifications to update the global vectors. It requires all vectors to be stored in the memory of all Spark executors, and, similarly to its single machine counterparts, is thus unsuitable for large vocabularies. The Deeplearning4j system takes a similar approach and thus suffers from the same limitations, although it does enable the use of GPUs to accelerate the training on each machine.

\subsection{Parameter servers}
A well-known distributed architecture for training very large machine learning models centers around the use of a parameter server to store the latest values of model parameters through the course of training.  A parameter server is a high performance, distributed, in-memory key-value store specialized to the machine learning training application.  It typically needs to support only fixed-size values corresponding to the model parameters, and also may support additive updates of values in addition to the usual key-value {\it get}s and {\it put}s.  A parameter server-based training system also includes a number of {\it worker}/{\it learner}/{\it client} nodes that actually carry out the bulk of the training computations.  The client nodes read in and parse training data in chunks or minibatches, fetch the model parameters that can be updated based on each minibatch, compute the updates (e.g., via gradient descent with respect to a minibatch restriction of the objective), and transmit the changes in parameter values to the parameter server shards which either overwrite or incrementally update these values in their respective in-memory stores.  As observed and partially theoretically justified in~\cite{hogwild} (see also~\cite{downpour}), in many applications involving sparse training data characterized by low average overlap between the model parameters associated with different minibatches, the model parameter updates arriving in parallel from multiple client nodes can be aggregated on the parameter server shards without locking, synchronization, or atomicity guarantees, and still result in a far better model accuracy versus training time latency trade-off than single threaded (i.e., sequential) training.

The parameter server paradigm has been applied successfully to the training of very large models for logistic regression, deep learning, and factorization machines, and to sampling from the posterior topic distribution in large-scale Latent Dirichlet Allocation~\cite{tensorflow,pslda1,downpour,difacto,scalingps,factorbird,swivel,pslda2,petuum}.   There have also been some attempts to extend the parameter-server approach to word2vec (e.g.,~\cite{dmtk}).  These have followed the above computational flow, with each parameter server shard storing the input and output vectors for a subset of the vocabulary.  Multiple client nodes process minibatches of the training corpus, determining for each word in each minibatch the associated context words and random negative examples, issuing {\bf get} requests to the parameter server shards for the corresponding vectors, computing the gradients with respect to each vector component, and issuing {\bf put} or {\bf increment} requests to update the corresponding vectors in the parameter server shards.

Unfortunately, such a conventional parameter server-based word2vec training system requires too much network bandwidth to achieve acceptable training throughput.   Using the skipgram training algorithm and denoting algorithm parameters as $ d $ for vector dimension, $ b $ for number of words per minibatch, $ w $ for average context size, and $ n $ for the number of random negative examples per context word, assuming negligible repetition of words within the minibatch and among the negative examples, and further assuming that vectors and their gradients are communicated and stored as arrays of single-precision floating point numbers at 4 bytes each, the amount of word vector data transferred for each {\bf get} and {\bf put} call from and to the parameter server, respectively, is on average
$ b \cdot (2 +  w \cdot n) \cdot d \cdot 4 $, or about
\begin{equation}
  r(w,n,d) = (2 + w \cdot  n) \cdot  d \cdot  4
  \label{eq:convbw}
\end{equation}
bytes per trained minibatch word.\footnote{This expression tends to lower bound the total bandwidth, as it accounts only for the word vector and gradient bytes.  The indices into the vocabulary sent with each {\bf get} and {\bf put} request require bandwidth as well, although this is small relative to the vector data.}  The formula arises from the fact that the input and output vectors for each term in the minibatch must be sent (this the '2' in the first factor in~(\ref{eq:convbw})), as must the output vectors for each random negative example.  There are on average $ w \cdot n $ of these per minibatch word.

For $ w = 10, n = 10, d = 500 $, values within the ranges recommended in~\cite{word2vec}, this works out to $ r(10,10,500) \approx 200,000 $ bytes transferred per word with each {\bf get} and {\bf put}. For 10 iterations of training on a data set of roughly 50 billion words, which is in the middle of the relevant range for the sponsored search application described in Section~\ref{sec:applications}, attaining a total training latency of one week using the above system would require an aggregate bandwidth of at least 1300Gbits/sec to and from the parameter servers\footnote{Obtained as $ 10\cdot 5e10 \cdot 2e5 \cdot 8 / (7 \cdot 24 \cdot  60 \cdot 60 \cdot 1e9) $.}. This is impractically large for a single application on a commodity-hardware shared compute cluster.  Moreover, one week training latency is already at the boundary of usefulness for our applications.

In the next section, we present a different distributed system architecture for word2vec that requires significantly less network bandwidth for a given training throughput than the above conventional parameter server-based system, while continuing to accommodate large vocabularies and providing sufficient computational power to achieve the higher throughput allowed by the reduction in network bandwidth.  



\section{Network-efficient distributed word2vec training system}  \label{sec:alg}
\begin{figure}
\centering
\includegraphics[width=3in]{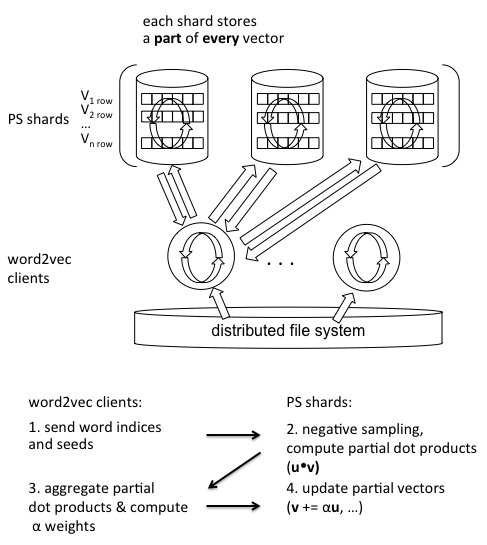}
\caption{Scalable word2vec on Hadoop.}
\label{fig:w2v}
\end{figure}
\subsection{Architecture}
Our distributed word2vec training system (i.e., for maximizing~(\ref{eq:w2vobjective})) is illustrated in Figure~\ref{fig:w2v}, with pseudo code for the overall computational flow in Figures~\ref{alg:w2v-server-dotprod}, \ref{alg:w2v-server-adjust}, and \ref{alg:w2v-client} in the Appendix.\footnote{Though we focus on the skipgram variant of word2vec, we note that the proposed approach readily extends to the continuous bag of words (CBOW) variant as well.}    As can be seen in Figure~\ref{fig:w2v}, the proposed system also features parameter-server-like components (denoted by ``PS shards'' in the figure), however they are utilized very differently and have very different capabilities from their counterparts in the conventional approach described above.   We shall, however, continue to refer to these components as parameter server shards.  The system features the following innovations, explained in more detail below, with respect to the conventional approach.
  \begin{itemize}
  \setlength{\itemsep}{2pt}
  \item Column-wise partitioning of word vectors among parameter server (PS) shards (as opposed to word-wise partitioning).
  \item No transmission of word vectors or vector gradients across the network.
  \item Server-side computation of vector dot products and vector linear combinations, distributed by column partitions.
  \item Distributed server-side generation of random negative examples via broadcasting of common random number generator seeds.
  \end{itemize}
In particular, avoiding the transmission of vectors and gradients greatly reduces network bandwidth requirements relative to the conventional approach.  We are not aware of any existing systems for training word2vec or its close relatives, matrix factorization and collaborative filtering (i.e., those systems cited in the previous section), that distribute vectors and compute in the manner of the proposed system.
  
In our system, a number of parameter server shards each stores a designated portion of {\em every} input (row) vector $ \bu(w) = [u_1,$ $ u_2,\ldots,$ $u_d] $ and output (row) vector $ \bv(w) = [v_1,$ $ v_2,$ $ \ldots,$ $v_d] $ (dependence of components on $ w $ is suppressed).  For example, assuming a vector dimension $ d =  300 $, 10 parameter server shards, and equi-partitioned vectors, shard $ s \in \{0,\ldots,9\} $ would store the 30 components of $ \bu(w) $ and $ \bv(w) $ with indices $ i $ in the range $ 30 s + 1 \leq i \leq 30s  + 30 $. We shall denote shard $ s $ stored portion of $ \bu(w) $ and $ \bv(w) $ as $ \bu_s(w) $ and $ \bv_s(w) $, respectively.  We refer to this as a 'column-wise' partitioning of the vectors, or more specifically, of the matrix whose rows correspond to the word vectors, as in 
 \[ \left[ \begin{array}{ccccc} \bu(w_1)^T & \bv(w_1)^T & \ldots & \bu(w_{|{\cal V}|})^T & \bv(w_{|{\cal V}|})^T \end{array} \right]^T \]
 where $ w_1,\ldots,w_{|{\cal V}|} $ are the words in the vocabulary according to a fixed ordering ${\cal O}$ (e.g.,\ by decreasing frequency of occurrence in the corpus).  In the sequel, we shall equate each word $ w_\ell $ with $ \ell $, its index in this ordering, so that $ \bu(w_\ell) \equiv \bu(\ell) $, and so on.
For $ S $ shards, the vocabulary size can thus be scaled up by as much as a factor of $ S $ relative to a single machine.

The vectors are initialized in the parameter server shards as in~\cite{word2vec}.
Multiple clients running on cluster nodes then read in different portions of the corpus and interact with the parameter server shards to carry out minibatch stochastic gradient descent (SGD) optimization of~(\ref{eq:w2vobjective}) over the word vectors, following the algorithm in Figure~\ref{alg:w2v-client} (in the appendix). Specifically, the corpus is partitioned into disjoint minibatches with index sets  $ {\cal B}_1, {\cal B}_2, \ldots, {\cal B}_N $ wherein each $ {\cal B}_h $ is a subset of (sentence index, word index) pairs.  For each $ {\cal B}_h $ the word vectors are adjusted based on the gradient of the summation~(\ref{eq:w2vobjective}) restricted to the input words belonging to $ {\cal B}_h $, as given by
\begin{multline}
  \Lambda({\cal B}_h) \stackrel{\triangle}{=} \sum_{(i,j) \in {\cal B}_h} \; 
\sum_{\substack{k \neq j: |k-j| \leq b_{i,j}, \\ w_{i,k} \in {\cal V}}} \Bigg[ \log \sigma (\bu(w_{i,j})\bv^T(w_{i,k})) + \\
\sum_{\tilde{w} \in {\cal N}_{i,j,k}}\log (1-\sigma (\bu(w_{i,j})\bv^T(\tilde{w}))) \Bigg].
  \label{eq:minibatch}
 \end{multline}

The gradient of $ \Lambda({\cal B}_h) $ with respect to the word vector components is 0 for all word vector components whose corresponding words do not appear as inputs, outputs, or negative examples in~(\ref{eq:minibatch}).  For the remaining components, the gradient is conveniently expressed in groups of components corresponding to specific word vectors.   For example, consider a pair of indices $(i_o,j_o)$ belonging to $ {\cal B}_h $. The gradient components corresponding to the word vector $ \bu(w_{i_o,j_o}) $ can be expressed as
\begin{multline}
\left. \grad \Lambda({\cal B}_h) \right|_{\bu(w_{i_o,j_o})} = \sum_{(i,j) \in {\cal B}_h: w_{i,j} = w_{i_o,j_o}} \sum_{\substack{k \neq j: |k-j| \leq b_{i,j}, \\ w_{i,k} \in {\cal V}}} \\ \Bigg[ (1- \sigma (\bu(w_{i_o,j_o})\bv^T(w_{i,k}))) \bv(w_{i,k}) - \\
    \sum_{\tilde{w} \in {\cal N}_{i,j,k}} \sigma (\bu(w_{i_o,j_o})\bv^T(\tilde{w}))\bv(\tilde{w}) \Bigg]
  \label{eq:minibatchgradex}
\end{multline}
We see that evaluation of $ \left. \grad \Lambda({\cal B}_h) \right|_{\bu(w_{i_o,j_o})} $ requires computing the dot (or inner) products $ \bu(w_{i_o,j_o})\bv^T(\cdot) $ appearing in the arguments to $ \sigma $ and then computing linear combinations of the vectors $ \{ \bv(w_{i,k}) \} $ and $ \{ \bv(\tilde{w}) \} $, with weights depending on the dot products.   A similar expression and computation applies to the other gradient components corresponding to other word vectors appearing in~$\Lambda({\cal B}_h) $.  The vector $ \bu(w_{i_o,j_o}) $ (and, correspondingly, the other vectors as well) are updated according to the usual SGD update rule
\begin{equation}
  \bu(w_{i_o,j_o}) \leftarrow \bu(w_{i_o,j_o}) + \alpha \left. \grad \Lambda({\cal B}_h) \right|_{\bu(w_{i_o,j_o})}
  \label{eq:gradupdate}
\end{equation}
where $ \alpha $ is a (suitably small) learning rate.

Once a client has assembled the indices (indexing according to the order $ {\cal O} $ above) of positive output examples and input words corresponding to a minibatch $ {\cal B}_h $, it interacts with the parameter server shards to compute~(\ref{eq:minibatchgradex}) and~(\ref{eq:gradupdate}) using two remote procedure calls (RPCs), {\bf dotprod} and {\bf adjust}, which are broadcasted to all PS shards, along with an intervening computation to aggregate results from the {\bf dotprod} RPC returned by each shard.  The RPC calls are detailed in Figures~\ref{alg:w2v-server-dotprod} and~\ref{alg:w2v-server-adjust} (in the Appendix), and, at a higher level, entail the following server/shard side operations:
\begin{itemize}
\item {\bf dotprod}: Select negative examples $ \tilde{w} $ in~(\ref{eq:minibatchgradex}) according to a probability distribution derived from the vocabulary histogram proposed in~\cite{word2vec}, but with the client thread supplied seed initializing the random number generation, and then 
  return all partial dot products required to evaluate the gradient~(\ref{eq:minibatchgradex}) for all positive output, negative output, and input word vectors associated with the minibatch, wherein the partial dot products involve those vector components stored on the designated shard: $ \bu_s\bv_s^T $. 
\item {\bf adjust}:  Regenerate negative examples used in preceding {\bf dotprod} call using the same seed that is again supplied by the client thread. Compute~(\ref{eq:gradupdate}) for vector components associated with the minibatch stored on the shard as a partial vector (restricted to components stored on shard) linear combination using weights received from the client.  
\end{itemize}
Between these two RPCs the client computes the linear combination weights needed for {\bf adjust} by summing the partial inner products returned by the shards in response to the {\bf dotprod} calls and evaluating the sigmoid function at values given by the aggregated dot products.  These weights are then passed to the {\bf adjust} RPC, along with the seeds for regenerating the identical random negative example indices $ \tilde{w} $ that were generated during the {\bf dotprod} RPC.  The retransmission simplifies the server in that state need not be maintained between corresponding {\bf dotprod} and {\bf adjust} calls.  Note that the same seeds are sent to all shards in both calls so that each shard generates the same set of negative example indices.  The shards are multithreaded and each thread handles the stream of RPC's coming from all client threads running on a single node.

In a typical at scale run of the algorithm, the above process is carried out by multiple client threads running on each of a few hundred nodes, all interacting with the PS shards in parallel.  The data set is iterated over multiple times and after each iteration, the learning rate $ \alpha $ is reduced in a manner similar to the open source implementation of~\cite{word2vec}.  Note that there is no locking or synchronization of the word vector state within or across shards or across client threads during any part of the computation.
The only synchronization in effect is that the RPC broadcast ensures that all shards operate on the same set of word vector indices for computing their portion of the corresponding calls.   Additionally, the client threads independently wait for all responses to their corresponding {\bf dotprod} calls before proceeding.  The lack of synchronization introduces many approximations into the overall SGD computation, similar in spirit to the HOGWILD\!~\cite{hogwild} and Downpour SGD~\cite{downpour} distributed optimization schemes.   For example, here, in the worst case, the state of the vectors associated with a minibatch could change between the {\bf dotprod} and {\bf adjust} calls issued by a single client thread. Nevertheless, despite such approximations, our distributed algorithm incurs surprisingly little degradation in the quality of the trained vectors as compared to single machine solutions (in cases where the computation can be carried out on one machine), as shown in Section~\ref{experiments}.

Two details of our version of the algorithm and implementation are helpful for improving convergence/performance on some data sets.  One is that in the {\bf adjust} computation (Figure~\ref{alg:w2v-server-adjust}) the word vectors belonging to the minibatch are not updated until the end of the call so that references to word vectors throughout the call are to their values at the start of the call.  The second is an option for interleaved minibatch formation, which can be used to ensure that indices $ (i,j) $ of input words belonging to a minibatch are sufficiently separated in the training corpus, and ideally, belong to different sentences.  This allows input word vectors within a sentence (which are linked through their overlapping output word windows) to ``learn'' from each other during a single training iteration, as their respective minibatches are processed. 

\subsection{Network bandwidth analysis}
Using the same notation as in~(\ref{eq:convbw}), and letting $ S $ denote the number of shards, the average bytes transferred from all PS shards for each {\bf dotprod} call is upper bounded by
\begin{equation}
  b \cdot (w \cdot (n + 1)) \cdot S \cdot 4.
  \label{eq:newbatchbw}
\end{equation}
That is, each shard transfers the partial dot product results between the input vector of each minibatch word and all context words (there are no more than an average of $ w $ of these per minibatch word) and negative examples (there are no more than $ n $ per context per minibatch word, or $ n \cdot w $ per minibatch word).

It is not hard to see that this is precisely the number of bytes transferred to all PS shards for the vector linear combination component of each {\bf adjust} call.  That is, there are two linear vector updates for each pair of vectors for which a dot product was computed, and these updates involve the same linear combination weight.  Normalizing (\ref{eq:newbatchbw}) by the minibatch size, we have the following counterpart of~(\ref{eq:convbw}) for the bytes transferred, in each direction, per trained minibatch word, for the proposed scheme:\footnote{Again, in this case, because the negative example indices are generated on the PS shards and not transmitted, the bandwidth for index transmission for $ n $ negative examples per context word can be seen to be $ 1/n $ the bandwidth of the partial dot products and linear combination weights, so that it is relatively small.}
\begin{equation}
  r'(w,n,S) = (w \cdot (n + 1)) \cdot S \cdot 4.
  \label{eq:newbw}
\end{equation}
Notice that the vector dimension $ d $ has been replaced by the number of shards $ S $.

The ratio of the network bandwidths of the proposed system and a conventional parameter server based system is
\[
  \frac{r'(w,n,S)}{r(w,n,d)} \approx \frac{S}{d}.
\]
For typical parameters of interest (we typically have $ S $ between 10 and 20, increasing with $ d $ between 300 and 1000), this is in the range of $ 1/20 $ to $ 1/100 $, effectively eliminating network bandwidth as a bottleneck for training latency, relative to the conventional approach.

\begin{table*}
\footnotesize
  \begin{center}
    \begin{tabular}{|l|l|l|l|}\hline
      test & single machine & distr.\ (low parallelism) & distr.\ (high parallelism) \\\hline
      phrase analogies accuracy & 0.73 & 0.72 & 0.70 \\
      wordsim 353 Spearman rank corr.& 0.66 & 0.69 & 0.67 \\\hline
    \end{tabular}
  \end{center}
  \caption{Word vector metrics on two semantic tests for various configurations.}
  \label{tab:accuracy}
\end{table*}

\vspace{-.25cm}
\section{Implementation on Hadoop}
\label{sec:implementation}
We have implemented the system described in Section~\ref{sec:alg} in Java and Scala on a Hadoop YARN scheduled cluster, leveraging Slider~\cite{slider} and Spark~\cite{spark}.   Our end-to-end implementation of training carries out four steps:  vocabulary generation, data set preprocessing, training, and vector export.   We next review the details of each of these steps.  Throughout, all data, including the initial training data, its preprocessed version, the exported vectors are all stored in the Hadoop Distributed File System (HDFS).  We remark that although our compute environment is currently based on Hadoop and Spark, other distributed computational frameworks such as the recently released TensorFlow could also serve as a platform for implementing the proposed system.\footnote{The demonstration word2vec systems in the latest TensorFlow release are single machine only.}  
\subsection{Main steps}
\subsubsection{Vocabulary generation} This step entails counting occurrences of all words in the training corpus and sorting them in order of decreasing occurrence.  As mentioned, the vocabulary is taken to be the $ V $ most frequently occurring words, that occur at least some number $ C $ times. It is implemented in Spark as a straight-forward map-reduce job.
\subsubsection{Preprocessing} In this step, each word in the training corpus is replaced by its index in the sorted vocabulary generated in the preceding phase (the ordering $ {\cal O} $ referred to in Section~\ref{sec:alg}).  This is also implemented in Spark using a low overhead in-memory key-value store to store the mapping from vocabulary words to their indices.  Our implementation hashes words to 64 bit keys to simplify the key-value store.
\subsubsection{Training} Referring to the system description in Section~\ref{sec:alg} (and Figure~\ref{fig:w2v}), the parameter server portion is implemented in Java, with the RPC layer based on the Netty client-server library~\cite{netty}.  The RPC layer of the client is implemented similarly.  The higher layers of the client (i/o, minibatch formation, partial dot product aggregation, linear combination weight computation) are implemented in Scala and Spark.  In particular, the clients are created and connect to the PS shards from within an RDD {\bf mapPartitions} method applied to the preprocessed data set that is converted to an RDD via the standard Spark file-to-RDD api.   At the start of training, the PS shards are launched from a gateway node onto Hadoop cluster nodes using the Apache Slider application that has been designed to launch arbitrary applications onto a Hadoop YARN scheduled cluster.  The IP addresses and ports of the respective PS shards are extracted and passed to the Spark executors (which in turn use them to connect respective clients to the PS shards) as a file via the standard spark-submit command line executed on a gateway node.  Each {\bf mapPartitions} operation in the clients is multi-threaded with a configurable number of threads handling the processing of the input data and the interaction with the PS shards.  These threads share the same connections with the PS shards.  The PS shards are also multi-threaded based on Netty, wherein a configurable number of worker threads process incoming {\bf dotprod} and {\bf adjust} requests from multiple connections in parallel.  Each shard has a connection to each Spark executor.  The word vector portions are stored in each PS shard in arrays of primitive floats, and as mentioned, their indices in the arrays coincide with the indices of their corresponding words in the vocabulary.   In the steady state, the PS allocates no new data structures to avoid garbage collection.  Objects are created only during start-up, and possibly during the fairly infrequent connection setups, as managed by the Netty RPC layer.

\subsubsection{Vector export}  In this final step, carried out after training has completed, the partial vectors stored in each PS shard are aggregated and joined with their respective words in the vocabulary and stored together as a text file in HDFS.  Again, we leverage Spark to carry out this operation in a distributed fashion, by creating an RDD from the vocabulary and using {\bf mapPartitions} to launch clients that {\bf get} the partial vectors from the PS shards for the respective partition of vocabulary words, combine the partial vectors and save corresponding word and vectors pairs to HDFS.


\subsection{Training step throughput}
\label{subsec:throughput}
To give an idea of the kind of training throughput we can achieve with this system, the following is one configuration we have used for training the sponsored search application on our Hadoop cluster:\footnote{2000+ nodes with 128GB memory, dual socket, 12 cores per socket, Intel Haswell (ES2680v3, 2.5GHz) servers; 10Gb/sec Ethernet}
\begin{itemize}
\setlength{\itemsep}{1pt}
\item {\bf Algorithm parameters:} 200 million word vocabulary,  5 negative examples, maximum of 10 window size
\item {\bf Training system parameters:}  200 Spark executors, 8 threads per spark executor, minibatch size of 200
\end{itemize}
yields the following training throughputs in minibatch input words per second (see Section~\ref{sec:word2vecopt} for the definition of input word), for varying PS shards and vector dimensions:  
\begin{center}
\footnotesize
\begin{tabular}{|l|l|l|}
\hline dim. & \# PS shards & throughput (input words/sec) \\\hline
300 & 15 & $1.6\times10^6$ \\
300 & 10 & $1.3\times10^6$ \\
300 & 6 & $1.0\times10^6$ \\
1000 & 25 & $ 1.2\times 10^6 $ \\\hline
\end{tabular}
\end{center}
For this data set and algorithm parameters, each input word has associated with it an average of about 20 positive context words and negative examples, so that the system is effectively updating about 21 times the third column in the table number of vectors per second.  For the first line of the table, for example, this is over 33 million 300 dimensional vector updates per second.   The conventional parameter server approach would require a total bandwidth of about 300 Gbps (30 server shards would be needed for this) to and from the parameter server for similar training throughput.   This is close to 10 percent of the fabric bandwidth in our production data center.   The proposed system requires only about 15 Gbps, making it far more practical for deployment to production in a shared data center, especially in light of the training latency for which this bandwidth must be sustained, which is about two days for data sets of interest.   Even more extreme is the last line of the table (the 1000 dim.\ case), for which the equivalent throughput conventional system would require 800 Gbps vs. 20 Gbps for the proposed system.

One important property of the training system is that its throughput at any given time is limited by the throughput of the slowest PS shard at that time.  With this in mind, we use the YARN scheduler resource reservation capability exported through Slider to minimize resource contention on {\em all} of the machines to which the PS shards are assigned, thereby achieving higher sustained throughput.  Another important property of the training system is that increasing the number of shards beyond some point is not helpful since the vector portions handled by each shard become so small that the random access memory transaction bandwidth (number of random cache lines per second) becomes the bottle neck.  This explains the limited throughput scaling with PS shards for the 300 dimensional case above.  Further optimization of the vector-store of each PS shard with respect to caching and non-uniform memory access might be beneficial.  We leave this for future investigation.
\vspace{-.25cm}
\section{Evaluation \& Deployment}
\label{experiments}
In this section, we provide evidence that the vectors trained by the proposed distributed system are of high quality, even with fairly aggressive parallelism during training. 
We also show bucket test results on live web search traffic that compare query-ad matching performance of our large-vocabulary model to the one trained using single-machine implementation, which led to the decision to deploy the proposed system in production in late 2015. 

\subsection{Benchmark data set}
\label{subsec:benchmark}
To compare the proposed distributed system we trained vectors on a publicly available data set collected and processed by the script 'demo-train-big-model-v1-compute-only.sh' from the open-source package of~\cite{word2vec}. This script collects a variety of publicly available text corpuses and processes them using the algorithm described in~\cite{word2vec} to coalesce sufficiently co-occurring words into phrases.  We then randomly shuffled the order of sentences (delimited by new line) in the data set, retaining order of words within each sentence.
The resulting data set has about 8 billion words and yields a vocabulary of about 7 million words and phrases (based on a cut off of 5 occurrences in the data set).  We evaluated accuracy on the phrase analogies in the 'question-phrases.txt' file and also evaluated Spearman's rank correlation with respect to the editorial evaluation of semantic relatedness of pairs of words in the well known wordsim-353 collection~\cite{wordsim353}.

The results are shown in Table~\ref{tab:accuracy}. The first column shows results for the single machine implementation of~\cite{word2vec}, the second for a 'low parallelism' configuration of our system using 50 Spark executors, minibatch size of 1, and 1 thread per executor, and the third column for a 'high parallelism' configuration again with 50 executors, but with minibatch size increased to 50 and 8 threads per executor. The various systems were run using the skipgram variant with 500 dimensional vectors, maximum window size of 20 (10 in each direction), 5 negative examples, subsample ratio of 1e-6 (see~\cite{word2vec}), initial learning rate of 0.01875, and 3 iterations over the data set.  It can be seen that the vectors trained by the 'high parallelism' configuration of the proposed system, which is the closest to the configurations required for acceptable training latency in the large-scale sponsored search application, suffers only a modest loss in quality as measured by these tests.  Note that this data set is more challenging for our system than the sponsored search data set, as it is less sparse and  there is on average more overlap between words in different minibatches.  In fact, if we attempt to increase the parallelism to 200 executors as was used for the training of the vectors described in the next subsection, training fails to converge altogether.  We are unsure why our system yields better results than the implementation of~\cite{word2vec} on the wordsim test, yet worse scores on the analogies test.   We also note that the analogies test scores reported here involve computing the closest vector for each analogy ``question'' over the entire vocabulary and not just over the 1M most frequent words, as in the script 'demo-train-big-model-v1-compute-only.sh' of~\cite{word2vec}.

\subsection{Sponsored Search data set}
\label{subsec:sponsored}
We conducted qualitative evaluation in the context of sponsored search application described in Section~\ref{sec:applications}.  Figure~\ref{fig:topa2q} shows the queries whose trained vectors were found to be most similar (out of 133M queries) to an example ad vector, along with the respective cosine similarities to the ad vector.   The figure shows the ten most and least similar among the 800 most similar queries, where we note that the ten least similar queries can still be considered to be fairly semantically similar. This particular set of vectors was trained for a vocabulary of 200M generalized words using the 300 dimensional vector, 15 PS shard settings described in Section~\ref{subsec:throughput}.
We found the vector quality demonstrated in Figure~\ref{fig:topa2q} to be the norm based on inspections of similar matchings of query vectors to a number of ad vectors.  

  We also compared the cosine similarities for pairs of vectors trained using the proposed distributed system and for corresponding vector pairs trained using the open--source implementation of~\cite{word2vec}, again on a large search session data set.  The former was trained using a vocabulary of 200 million generalized words while the latter was trained using about 90 million words which is the most that could fit onto a specialized large memory machine.   
 For a set of 7,560 generalized word pairs with words common to the vocabularies trained by the respective systems we found very good agreement in cosine similarities between the corresponding vectors from the two systems, with over 50\% of word pairs having cosine similarity differences less than 0.06, and 91\% of word pairs having differences less than 0.1.



\begin{figure}
  \begin{center}
  \fbox{
    \parbox{3in}{
\footnotesize
{\bf ad title}: Download Piano Sheet Music \\
{\bf ad description}: World's Largest Selection of Sheet Music for Piano. Shop Now! \\

piano\_sheet\_music\_silent\_night, 0.963 \\
letter\_notes\_for\_songs, 0.960 \\
piano\_sheet\_music\_with\_lyrics, 0.955 \\
easy\_songs\_for\_the\_piano, 0.955 \\
easy\_music\_to\_play\_on\_the\_piano, 0.954 \\
super\_easy\_piano\_music, 0.954 \\
easy\_piano\_fur\_elise\_sheet\_music, 0.953 \\
easy\_piano\_notes\_for\_songs, 0.953 \\
sheet\_music\_for\_easy\_piano\_songs, 0.953 \\
\underline{easy\_piano\_songs\_sheet\_music, 0.953 \hspace{1in}} \\
free\_piano\_songs, 0.924 \\
free\_sheet\_music\_on\_line, 0.924 \\
a\_thousand\_years\_piano\_sheet\_music\_free, 0.924 \\
sheet\_music\_the\_lion\_sleeps\_tonight, 0.924 \\
music\_notes\_for\_free, 0.924 \\
hiawatha\_rag\_sheet\_music\_free, 0.924 \\
oceans\_piano\_sheet\_music, 0.924 \\
i\_have\_returned\_sheet\_music, 0.924 \\
free\_sheet\_music\_for\_vocal\_solos, 0.924 \\
piano\_chopsticks\_sheet\_music, 0.924
}
  }
\end{center}
  \caption{The top 10 and bottom 10 among the 800 most similar queries to a given ad vector, with cosine similarities to the ad vector.} 
\label{fig:topa2q}
\end{figure}

\begin{table*}
\footnotesize
  \begin{center}
    \begin{tabular}{|l|l|l|l|l|l|}\hline
      Bucket test & Query Coverage & Auction Depth & CTR & Click Yield & Revenue per Search \\\hline
      single machine training & +1.14\% & +2.13\% & +0.5\% & +1.70\% &  +7.07\%  \\
distributed training & +2.44\% & +2.39\%  & +0.2\% & +1.81\% &  +9.39\% \\\hline
    \end{tabular}
  \end{center}
  \caption{Comparison of broad match methods in A/B test.}
  \label{tab:abtest}
\end{table*}

\subsection{Online A/B tests}
Following successful offline evaluation of the proposed distributed system, in the following set of experiments we conducted tests on live web search traffic. We ran two bucket tests, each on $5\%$ of search traffic, where we compared query-ad matches produced by training query and ad vectors using search session data set spanning $9$ months of search data. One model was trained using implementation from \cite{word2vec} and the other was trained using the proposed distributed system. Both buckets were compared against control bucket, which employed a collection of different broad match techniques used in production at the time of the test. Each of the online tests were run for 10 days, one after another, more than a month apart. The results of the tests were reported in terms of query coverage (portion of queries for which ads were shown), Auction Depth (number of ads per query that made it into an auction) click-through rate (CTR, or number of ad clicks divided by number of ad impressions), click yield (number of clicks), and revenue. Instead of the actual numbers we show relative improvement over control metrics.

Both methods produced a separate query-ad match dictionary by finding $K=30$ nearest ads in the embedding space for each search query from our vocabulary, and keeping only ads with cosine similarity above $\tau=0.65$. The threshold was chosen based on editorial results. To implement the bucket test the query-ad match dictionary is produced offline and cached in the ad server memory such that ads can be retrieved in real-time given an input query. Post retrieval, a click model is used to estimate the clickability of the ad for that query and the ad is sent into an auction, where it competes with ads retrieved by other broad match algorithms. It gets to be shown to the user in case it wins one of the ad slots on the page. 

The first A/B test was conducted to evaluate the value of query-ad dictionary produced by single-machine implementation. This implementation could scale up to a model with $50$M query vectors. It was compared against control bucket that ran a production broad match module. Following positive A/B test metrics, with improvements in coverage and revenue, presented in the first row of Table \ref{tab:abtest}, the dictionary was launched to production and incorporated into the existing broad match production model.

The second A/B test was conducted to evaluate incremental improvement over the single machine solution, which was already launched in production. The model contained vectors for $133$M queries. As it can be observed in the second row of Table \ref{tab:abtest}, the distributed solution provided additional 2.44\% query coverage and additional 9.39\% revenue, without degrading user experience (CTR remained neutral).

This strong monetization potential of our distributed system for training large vocabularies of query and ad vectors led to its deployment in our sponsored search platform. The model is being retrained on a weekly basis, automated via Apache Oozie\cite{oozie}, and is currently serving more than $30\%$ of all broad matches. 

\section{Conclusion}  \label{sec:conclusion} 
In this paper, we presented a novel scalable word2vec training system that, unlike available systems, can train semantically accurate vectors for hundreds of millions of vocabulary words with training latency and network bandwidth usage suitable for regular training on commodity clusters.   
We motivated the usefulness of large vocabulary word2vec training with a sponsored search application involving generalized ``words'' corresponding to queries, ads, and hyperlinks, for which the proposed system has been deployed to production. The results on both benchmark data sets and online A/B tests strongly indicate the benefits of the proposed approach.



\begin{appendix}
\begin{algorithm}[ht]
\footnotesize
$PS_s$.\textbf{dotprod}($W_{\rm input}$, $ W_{\rm output}$, long $seed$) \\
$ R \leftarrow $ Random Number Generator initialized with $seed$ \;
  $ pos = 1$; $neg = 1 $ \;
\tcc{iterate over words in minibatch}
\For{$i \leftarrow 1$ \KwTo $|W_{\rm input}| $}{
  \tcc{iterate over words in context}
  \For{$j \leftarrow 1$ \KwTo $|W_{\rm output}[i]|$ }{
    $ w_I \leftarrow W_{\rm input}[i] $;  $ w_O \leftarrow W_{\rm output}[i][j] $ \;
    \tcc{generate $ N $ random negative examples for current output word}
    $ NS \leftarrow $ Array($ N $ negative word indices $ \neq w_O $, generated using $R$) \;
    \tcc{compute partial dot products for positive and negative examples}
      $ F^+[pos{+}{+}] =  \bu_s(w_I)\bv_s^T(w_O) $ \;
      \For{ $ ns  \leftarrow NS $ }{
         $ F^-[neg{+}{+}] = \bu_s(w_I)\bv_s^T(ns) $ \;
      }
  }
}
\tcc{send results back to client}
\Return $ (F^+_s, F^-_s) $
 \caption{Server side computation - {\bf dotprod}.}
 \label{alg:w2v-server-dotprod}
\end{algorithm}
\begin{algorithm}[ht]
\footnotesize
void $PS_s$.\textbf{adjust}($W_{\rm input}$, $ W_{{\rm output}}$, $G^+$, $G^-$, $seed$) \\
$ R \leftarrow $ Random Number Generator initialized with $seed$ \;
$ pos = 1$; $neg = 1 $; $ \bdu_s(\cdot) = 0$; $\bdv_s(\cdot) = 0$\;
\For{$i \leftarrow 1$ \KwTo $|W_{\rm input}| $}{
  \For{$j \leftarrow 1$ \KwTo $|W_{{\rm output}}[i]|$ }{
    $ w_I \leftarrow W_{\rm input}[i] $; $ w_O \leftarrow W_{\rm output}[i][j] $ \;
    \tcc{regenerate random negative examples}
    $ NS \leftarrow $ Array($ N $ negative word indices $ \neq w_O $, generated using $R$) \;
    \tcc{compute partial gradient updates and store in scratch area}
      $ \bdu_s(w_I) {+}{=} G^+[pos]\bv_s(w_O) $; $ \bdv_s(w_O) {+}{=} G^+[pos{+}{+}]\bu_s(w_I) $ \;
      \For{ $ ns  \leftarrow NS $ }{
         $ \bdu_s(w_I) {+}{=} G^-[neg]\bv_s(ns) $; $ \bdv_s(ns) {+}{=} G^-[neg{+}{+}]\bu_s(w_I) $ \;
      }
  }
}
\tcc{add partial gradient updates to partial vectors in store}
\For{all $w$}{
  $\bu_s(w){+}{=}\bdu_s(w)$; $ \bv_s(w){+}{=}\bdv_s(w) $
}
  \caption{Server side computation - {\bf adjust}.}
  \label{alg:w2v-server-adjust}
\end{algorithm}

\begin{algorithm}[ht]
\footnotesize
  \SetKwInOut{Input}{input}\SetKwInOut{Output}{output}
 \Input{${\cal V}$: Vocabulary, $ \{ {\cal P}_i \} $: training data partitions }
 \Output{$\bu_{i}$: Vectors for vocabulary words}
$S$ = \# of parameter servers needed for $|{\cal V}|$ words \;
Launch parameter servers $ \{PS_1,\ldots, PS_S\} $ \;  
Initialize vectors in PS server \;
\For{iteration $\leftarrow$ $1,\ldots,\# Iterations$}{
  UnprocessedPartitions  $ \leftarrow $ $ \{ {\cal P}_i \} $ \;
  \For{each executor, in parallel}{
    \While{UnprocessedPartitions is non-empty}{
          $ p $ $ \leftarrow $ next partition in UnprocessedPartitions
      Launch client $cl$ connected to $\{PS_j\} $ \;
    \For{$ {\cal B} $ $ \leftarrow $ minibatches in $ p $}{
      $seed$ = randomly select a seed \;
      $W_{\rm input}[]$ $ \leftarrow $ Array of word indices in $ {\cal B} $\;
      $W_{\rm output}[][]$ $ \leftarrow $ Array of Arrays of context word indices of words in $ {\cal B} $ \;
       \scriptsize
      \tcc{client broadcasts word indices to shards which compute partial dot products in parallel, returning results to client}
       \footnotesize
   \For{$s \leftarrow 1 $ \KwTo $ S $, in parallel}{
     $(F^+_s,F^-_s)$ = $PS_s$.{\bf dotprod}($W_{\rm input}$, $W_{\rm output}$, $seed$)\;
   }
   \scriptsize
   \tcc{aggregate partial dot products and compute linear coefficients for gradient update}
   \footnotesize
   $ (F^+,F^-) \leftarrow \sum_s(F^+_s,F^-_s) $ \;
   $G^+ \leftarrow \alpha(1 - \sigma(F^+)) $ ; $G^- \leftarrow -\alpha\sigma(F^-) $ \;
   \scriptsize
   \tcc{client broadcasts coefficients to shards which compute partial vector linear combinations}
   \footnotesize
   \For{$s \leftarrow 1 $ \KwTo $ S $, in parallel}{
     $PS_s$.{\bf adjust}($W_{\rm input}$, $W_{\rm output}$, $G^+$, $G^-$, $seed$)\;
     }
   }
    }
  }
  }
\Return input vectors $\{\bu$\} from $ \{PS_1, ..., PS_{S}\} $\;
 \caption{Grid based word2vec algorithm.}
\label{alg:w2v-client}
\end{algorithm}

\end{appendix}
\end{document}